\documentclass[reprint, cha]{revtex4-1}
\usepackage{amssymb}
\usepackage{graphicx}
\usepackage{amsmath}
\usepackage{amsfonts}
\usepackage{amssymb}
\usepackage{amsthm}
\usepackage{verbatim}
\usepackage{enumerate}
\usepackage{natbib}
\usepackage{mdwlist} 
\usepackage{subfigure}
\usepackage{color}
\usepackage[utf8]{inputenc}

\usepackage{algpseudocode}
\usepackage{algorithmicx}
\usepackage{lineno}

\begin{document}

\title{Fast deterministic tourist walk for texture analysis}

\author{Lucas Correia Ribas}
 	     \email{lucasribas@usp.br}
\affiliation{Institute of Mathematics and Computer Science, University of S\~{a}o Paulo (USP), Avenida Trabalhador s\~{a}o-carlense, 400 13566-590 S\~{a}o Carlos, S\~{a}o Paulo, Brazil} 
\affiliation{Scientific Computing Group, S\~ao Carlos Institute of Physics, University of S\~{a}o Paulo (USP),  PO box 369 13560-970 S\~{a}o Carlos, S\~{a}o Paulo, Brazil - www.scg.ifsc.usp.br}

\author{Odemir Martinez Bruno}
              \email{bruno@ifsc.usp.br}
\affiliation{Scientific Computing Group, S\~ao Carlos Institute of Physics, University of S\~{a}o Paulo (USP),  PO box 369 13560-970 S\~{a}o Carlos, S\~{a}o Paulo, Brazil - www.scg.ifsc.usp.br}

\date{\today}

\begin{abstract}

Deterministic tourist walk (DTW) has attracted increasing interest in computer vision. In the last years, different methods for analysis of dynamic and static textures were proposed. So far, all works based on the DTW for texture analysis use all image pixels as initial point of a walk. 
However, this requires much runtime.
In this paper, we conducted a study to verify the performance of the DTW method according to the number of initial points to start a walk.
The proposed method assigns a unique code to each image pixel, then, the pixels whose code is not divisible by a given $k$ value are ignored as initial points of walks. Feature vectors were extracted and a classification process was performed for different percentages of initial points. Experimental results on the Brodatz and Vistex datasets indicate that to use fewer pixels as initial points significantly improves the runtime compared to use all image pixels. In addition, the correct classification rate decreases very little.
\end{abstract}


\maketitle

\section{Introduction} 
\label{sec:intro}
Deterministic walk on random and regular means have achieved promising results over the years with highlight for deterministic tourist walk (DTW). The initial study of the DTW was introduced by Lima, Martinez and Kinouchi \cite{PhysRevLett.87.010603}, with the goal to analyze the effect of simple walks on random environments. These walks are automata able to obtain several characteristic or rich information about the environment on that are performed. Here, we are interested in the deterministic tourist walk algorithm for texture analysis.

Textures can be defined as a repetition of elements on a surface with visual patterns composed of sub-patterns with brightness, color, shape and size. Also, it is an important visual attribute widely used to describe patterns and characteristics of images in computer vision and image processing \cite{Goncalves20132953,Backes20111684,Ribas2015,Backes:2010:TAB:1840008.1840249,xu2009viewpoint}.
Recently, the deterministic tourist walk has emerged as a very promising approach for texture analysis \cite{backesTourist,backes2010texture,Backes20111684,Goncalves201211818,Goncalves20131163,Goncalves20134283,Goncalves2011}.  The deterministic tourist walk can be understood as a tourist that visits cities in a map according to a rule of movement and a memory. In images, each pixel is considered a city and the neighborhood of a pixel is the 8-connected. A walk is started from each pixel and the tourist moves according to the following rule: goes to nearest pixel (i.e., that minimizes the module of the difference between the gray levels of the pixels) that has not been visited in the last $u$ steps. This walk produces a trajectory composed by two parts: transient and attractor. The transient is the initial part of the walk and the attractor, the final part, it is a cycle of cities that the tourist gets stuck.

The precursor study for texture analysis using DTW was introduced in Ref. \cite{campiteliTourist}. This study uses the transient and attractor sizes for texture characterization. Then, Backes et al. \cite{backes2010texture} proposes a new variation that combines two rules of movement and new measures extracted from the joint distribution of attractor and transient sizes.
After that, a new approach combining DTW and fractal dimension to extract spatial features of attractors achieved a significant improves in classification~\cite{Goncalves20132953}. Besides these, other works investigated the use of the DTW combined with complex networks for texture analysis \cite{Goncalves201211818,Backes20111684}. Another field with good results are the dynamic textures. These are an extension of the textures for the spatial-temporal domain, i.e., texture patterns in videos. In Refs. \cite{Goncalves20134283,Goncalves20131163}, were proposed two methods for dynamic texture recognition and segmentation with significant improvement in the field.

For image characterization, all works cited above use all image pixels as initial points to perform a deterministic tourist walk. Thus, the $N$ image pixels are considered the map of the tourist. However, although the several studies performed, there is not a full investigation that analyzes the ideal number of initial points and its influence in the characterization. From the first work, it was assumed all image pixels as the map, without analyzing the consequences. It is certain that this strategy increases considerably the computational cost of the methods that use the DTW. Therefore, this is a point that need be investigated more deeply.

This paper accomplishes a study to analyze the performance of the deterministic tourist walk method according to the number of initial points to start the walks. The main objective is to verify if the DTW method is competitive using a smaller number of initial points. For this, we propose a simple method to select the pixels in that starts a walk. This method is deterministic and selects pixels of all regions from the image. Thus, three characteristics important for us are considered: homogeneity, simplicity and deterministic. From this characteristic, it is expected that our method has good results with a low computational cost. Experimental results using textures from the Brodatz book and Vistex datasets illustrate that our method improves considerably the computational cost and achieves good correct classification rate compared with the original method. Thus, it is possible  to obtain a competitive running time, without significantly decrease the correct classification rate. The main contribution of this paper is a simple and effective optimization of the deterministic tourist walk that can be used in all methods of dynamic and static textures of the literature.

The rest of the paper is structured as follows. In Section \ref{sec:turista}, the deterministic tourist walk method is described. The proposed method of the fast deterministic tourist walks is presented in Section \ref{sec:fast}. In Section \ref{sec:exp}, we describe the datasets and experimental  setup used to evaluate the proposed method. Section \ref{sec:results} presents the results and discussion. Finally, the Section \ref{sec:conc} concludes the paper and presents suggestions for further investigation.
\section{Deterministic tourist walk}\label{sec:turista}

To describe the deterministic tourist walk on image, consider which an image consists of a pair $(P,I)$, where $P$ is finite set of pixels and $I$ a mapping that assigns to each pixel $i=(x_i,y_i)$ an intensity $I(i) \in [0,255]$.
The neighborhood $\eta(i)$ consists of pixels $j$ whose the Euclidean distance between $i$ and $j$ is less than $\sqrt{2}$ (Equation \ref{eq:dist}).

\begin{equation}
\begin{aligned}
\eta(i) = \{j \mid d(i,j) \leq \sqrt{2}\}, \\
d(i,j)=\sqrt{(x_i-x_j)^2 + (y_i-y_j)^2}
\label{eq:dist}
\end{aligned}
\end{equation} 

The "distance" between two pixels $i$ and $j$ neighbors is given by a weight $w(i,j)$. This weight is the modulus of difference between their intensities:

\begin{equation}
w(i,j)= \mid I(i) - I(j) \mid
\label{eq:peso}
\end{equation}

Given the above definitions, for texture characterization is considered independent walks started from each pixel $i \in P$.
The tourist walks through image pixels according to a rule of movement. Considering that a tourist is in the pixel $i$, the next step is move to one of neighboring pixels $j$ that minimizes the weight $w(i,j)$.  Moreover, the tourist avoids returning to pixels visited in the last $\mu$ previous steps, which are stored in a memory $M$.  This rule of movement will be referred to as $r = min$. The movement of the tourist can also be governed by a rule of movement $r=max$. In this rule, the tourist walks towards the maximum weight $w(i,j)$. Another common situation is the existence of two or more neighbors that satisfy the rule of movement. In this case, the first neighbor is selected, considering which the neighbors are visited in a clockwise order \cite{backes2010texture}.  These rules produce trajectories of great complexity \cite{Backes20111684}.

\begin{figure}[!htbp]
	\centering
	\includegraphics[width=0.45\textwidth]{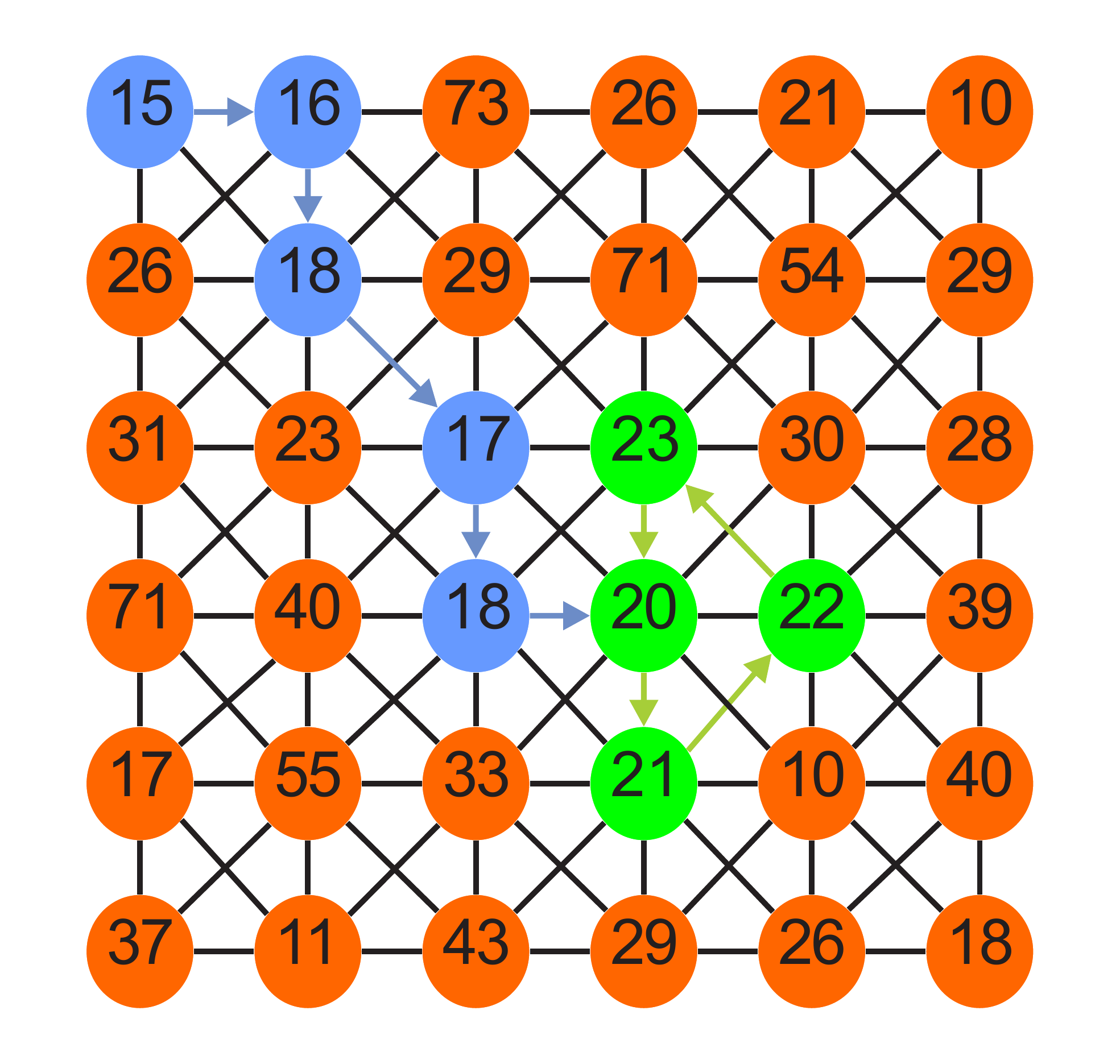}		
	\caption{Example of the tourist walk on image with memory $u=2$ and rule of movement $r=min$.}
	\label{fig:turist_image}
\end{figure}

Trajectories yielded from these walks can be divided into two parts: an initial part, with $\tau$ steps called transient and the final part, where the tourist is trapped in a cycle of period $\rho \leq \mu +1$ called attractor \cite{Goncalves20134283}. The attractor consists of a set of pixels which form a path that the tourist cannot escape from. There are cases where the tourist cannot find an attractor. In this case, the walk is performed until it finds a transient with a size equal to the number of image pixels $T$ \cite{backes2010texture}. In Figure \ref{fig:turist_image}, we show an example of the deterministic tourist walk on the image. In this example, the pixels are represented by circles and the gray-level from each pixel is given inside its circle. The transient is represented by the blue circles ($\tau=5$) and the attractor part is given by green circles ($\rho=4$).

For each initial situation, the tourist produces a different trajectory. Thus, according to the literature methods each pixel is taken as an initial condition for a trajectory. To obtain features able to discriminate texture images, the transient part and the attractor from each trajectory can be used to built a joint distribution $S_{\mu}^{r}(t,\rho)$ \cite{Goncalves201211818,Goncalves20131163,backes2010texture}. This joint distribution defines the probability of trajectories with transient $\tau$ and attractor $\rho$ sizes (Equation \ref{eq:dist_walk}).

\begin{equation}\label{eq:dist_walk}
S_{\mu}^{r}(\tau,\rho) = \frac{1}{N} \sum_{i=1}^{N} 
\left\{\begin{matrix}
1, & \textnormal{if } \tau_i = \tau \textnormal{ and } \rho_i = \rho \\
0, & \textnormal{otherwise}
\end{matrix}\right.
\end{equation}

where $i$ is the pixel that the trajectory was initiated and $N$ the number of initial points (pixels).

\section{Fast deterministic tourist walk}
\label{sec:fast}

As described above, a walk is started from each image pixel. It causes the literature methods have a high computational cost. Therefore, we propose a simple optimization for reducing the computational cost in texture image characterization. To describe our method, let us consider a mapping unidimensional for each pixel $i \in P$. Thus, to each pixel $i$ is assigned a code $c_i$ (Equation \ref{eq:code}).

\begin{equation}\label{eq:code}
c_i = W*x_i + y_i,
\end{equation}
where $W$ is the width of the image and, $x_i$ and $y_i$ the coordinates of the pixel $i$.

To optimize the deterministic tourist walk, some image pixels are ignored as initial point. For this, a simple way is apply a function $\delta(P,k)$ to the original set of pixels P, thus selecting a subset $P_k, P_k \subseteq P$, where each pixel of $i \in P_k$ has code $c_i$ do not divisible by $k$ (Equation \ref{eq:rest}). This is, if the remainder of the Euclidean division of $c_i$ by $k$ is different to 0. This approach enables us reducing the computational cost keeping the selecting of the pixels deterministic and homogeneously. It is important to stress that the choice of the pixel where starts a walk has that be homogeneous, i.e., pixels of all regions of the image will be chosen. Thus, it is possible to ensure that features of all regions of the image will be extracted by the method. 

\begin{equation}\label{eq:rest}
P_k = \delta(P,k) = \bigcup_{i \in P} i \mid \textnormal{if } c_i\pmod k \neq 0
\end{equation}

\begin{figure*}[!htbp]
	\centering
	\includegraphics[width=1\textwidth]{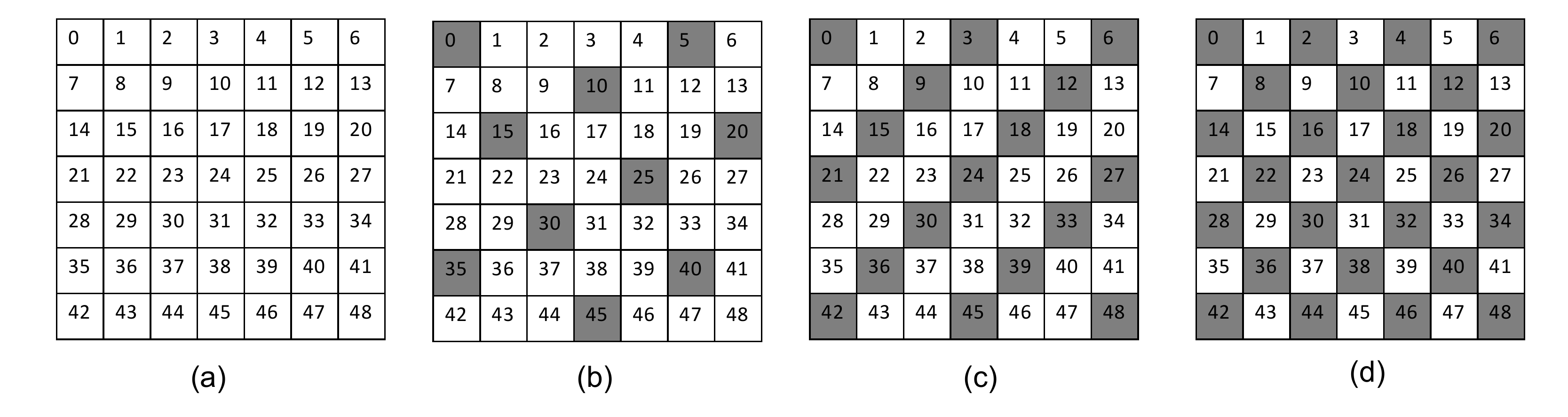}		
	\caption{Example of selection of initial points of an image with different $k$ values. Each pixel of the image is represented by a square and the code $c_i$ is given inside it. The white squares are initial points selected and the gray squares are initial points ignored. The $k$ values are: (a) original image, (b) $k=5$, (c) $k=3$ and (d) $k=2$. }
	\label{fig:example_fast}
\end{figure*}

Examples of application of the function of selection $\delta(P,k)$ on the image with various $k$ values can be seen in Figure \ref{fig:example_fast}. In the Figure, the white squares represent pixels selected (which belongs to $P_k$) to perform the walk and the gray squares are pixels ignored. Note that to increment $k$, the number of initial points decreases. Therefore, the $k$ value is directly associated to percentage of initial points which it will be performed walks in the image. Table \ref{tab:pct} shows the $k$ values and its percentage corresponding of initial points in the image.

\begin{table*}[!t]
	\caption{Percentage of initial points according to the $k$ values compared to the number original of image pixels.}
	\label{tab:pct}
	\centering
	\begin{tabular}{|c|c|c|c|c|c|c|c|c|c|}
		\hline
		$k$&10 & 7 & 5 & 4 & 3 & 2 & [2 9] & [2 5] & [2 3]\\
		\hline
		\% of initial points & 90\% & 86 \% &  80 \% &  75 \% &  67 \% &  50 \% &  44 \% &	 40 \% & 33 \%\\
		\hline
	\end{tabular}
\end{table*}

From this new set of pixels $P_k$, walks are performed to characterize the texture image.
Thus, a new joint distribution $S_{\mu,r}^{n}(\tau,\rho)$ can be obtained from the set $P_k$:

\begin{equation}\label{eq:dist_walk2}
S_{\mu,r}^{k}(\tau,\rho)= \frac{1}{\rvert P_k \lvert} \sum_{i \in P_k} 
\left\{\begin{matrix}
1, & \textnormal{if } \tau_i = \tau \textnormal{ and } \rho_i = \rho \\
0, & \textnormal{otherwise}
\end{matrix}\right.
\end{equation}

\subsection{Feature vector}

Several studies in previous works were performed to feature extraction from the joint distribution \cite{Goncalves201211818,backes2010texture,Goncalves20131163,Goncalves20134283}.
Here, we use a histogram $h_{\mu,r}^n(l)$, where $l = t+\rho$ (Equation \ref{eq:hist}). This histogram proved to achieve better results in classification tasks. The histogram $h_{\mu,r}^n(n)$ computes the frequency of trajectories that have size equal to $l=t + \rho$ in the joint distribution. 

\begin{equation}\label{eq:hist}
h_{\mu,r}^{k}(l)= \sum_{b=0}^{l-1} S_{\mu,r}^{k}(b,l-b)
\end{equation}

Earlier works have shown that the histogram concentrates most image information on the first elements \cite{backes2010texture,Goncalves201211818}. As there are no attractors of size smaller than $\mu+1$, the first position of the histogram used as descriptor is to $\mu+1$.
To compose the feature vector $\nu_{\mu}^r$, the first $m=4$ elements are obtained from the histogram.
Thus, the feature vector $\nu_{\mu}^r$ is constructed for a specific $\mu$ value:

\begin{equation}\label{eq:vector}
\nu_{\mu}^r = [h_{\mu,r}^{k}(\mu+1), h_{\mu,r}^{k}(\mu+2),...,h_{\mu,r}^{k}(\mu+m) ].
\end{equation}

Different rules of movement and values of memory size can obtain local and global informations of the image (e.g., low values of $\mu$ perform a better local analysis in the image) \cite{Goncalves20134283}. Therefore, different values of memory size and rule of movement are used to provide more effective texture representation. The final feature vector consists of the concatenation of $\nu_{\mu}^r$ vectors, for different $\mu$ values:
\begin{equation}\label{eq:final_vector}
\varphi^r = [\nu_{\mu_1}^r , \nu_{\mu_2}^r ,..., \nu_{\mu_n}^r ]
\end{equation}

where $\mu$ is the memory size and $r$ the rule of movement that can be $max$ or $min$.

\section{Experimental setup} \label{sec:exp}

\begin{figure}[h!]
	\centering
	\includegraphics[width=.45\textwidth]{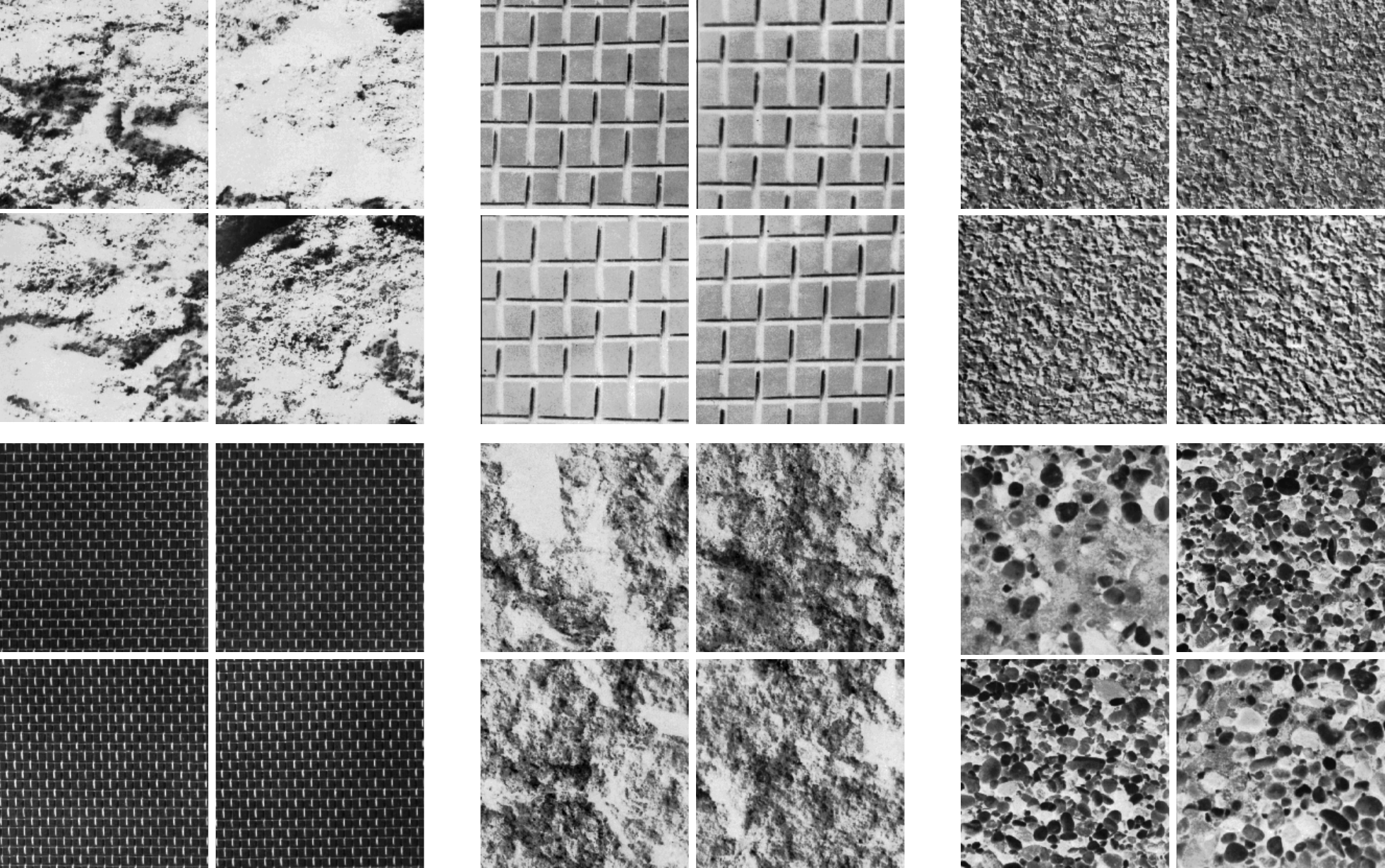}		
	\caption{Six examples of textures from the Brodatz dataset.}
	\label{fig:brodatz}
\end{figure}

In order to validate the proposed method and compare its efficiency with other ones in texture recognition, experiments were carried out using two datasets: (i) Brodatz and (ii) Vistex.
The Brodatz dataset consists of synthetic images collected from the Brodatz book \cite{brodatz1966textures}. This dataset is widely used in computer vision literature as a benchmark for evaluating methods of texture recognition. Each image has 200 $\times$ 200 pixels size with 256 gray levels. A total of 110 classes, with 10 samples each, were used. Samples of the textures are shown in Figure \ref{fig:brodatz}.

\begin{figure}[h!]
	\centering
	\includegraphics[width=.45\textwidth]{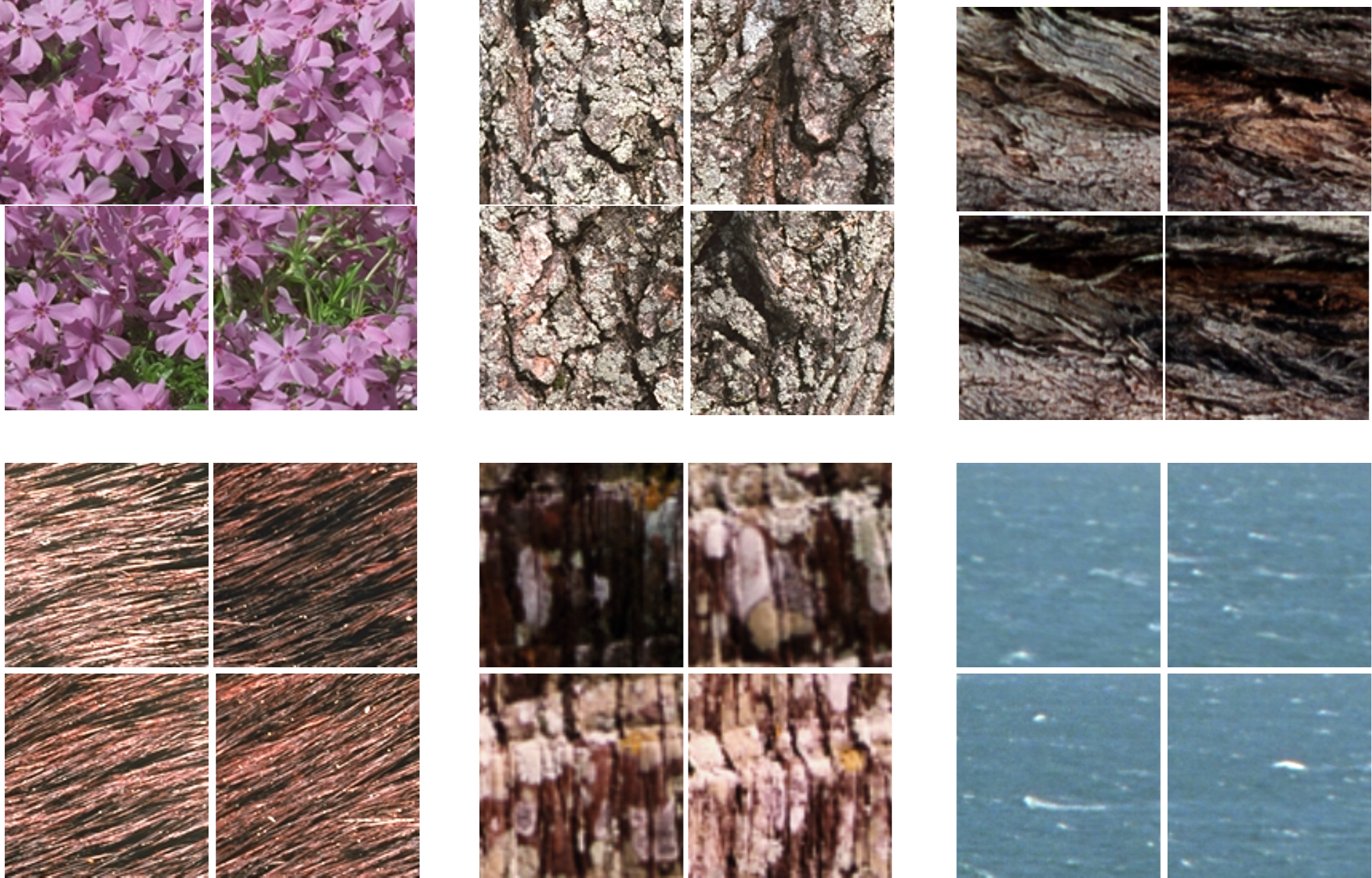}		
	\caption{Examples of some Vistex texture classes used in the experiments.}
	\label{fig:vistex}
\end{figure}

The second dataset used in this work, namely Vision Texture - Vistex \cite{picard1995vision}, contains 864 samples of real-world textures.
This dataset contains textures captured in different combinations of illumination, orientation and plane perspectives.
Each image has 128 $\times$ 128 pixels with 256 gray levels.
A total of 54 texture classes, each containing 16 samples, were used in the experiments.
Figure \ref{fig:vistex} shows some examples of textures.

Feature vectors were classified using Linear Discriminant Analysis (LDA) \cite{everitt1993principal} in a 10-fold cross-validation scheme.
LDA is a well-known method and supervised method to estimate a linear subspace with good discriminative properties. 
This method consists in to find data where the variance inter-classes is large in comparison to the variance intra-classes. 
The method of texture is considered good when it creates compact clusters far away from each other for all classes. 
The 10-fold cross-validation scheme splits the samples into 10 folds with the same number of sample per fold. One fold is used for test, while the folds remaining are used for supervision training and validation tasks. This process is performed for all folds.


\section{Results and discussion} \label{sec:results}

Here, first, we present an analyze of the parameters of the deterministic tourist walk and its impact on the texture recognition performance.
Then, we show the performance of the fast deterministic tourist walk for different $k$ values.

\subsection{Parameter evaluation}
\begin{figure}[h!]
	\centering
	\subfigure[Different memory sizes $\mu$]{\label{fig:memoryr} \includegraphics[width=0.5\textwidth]{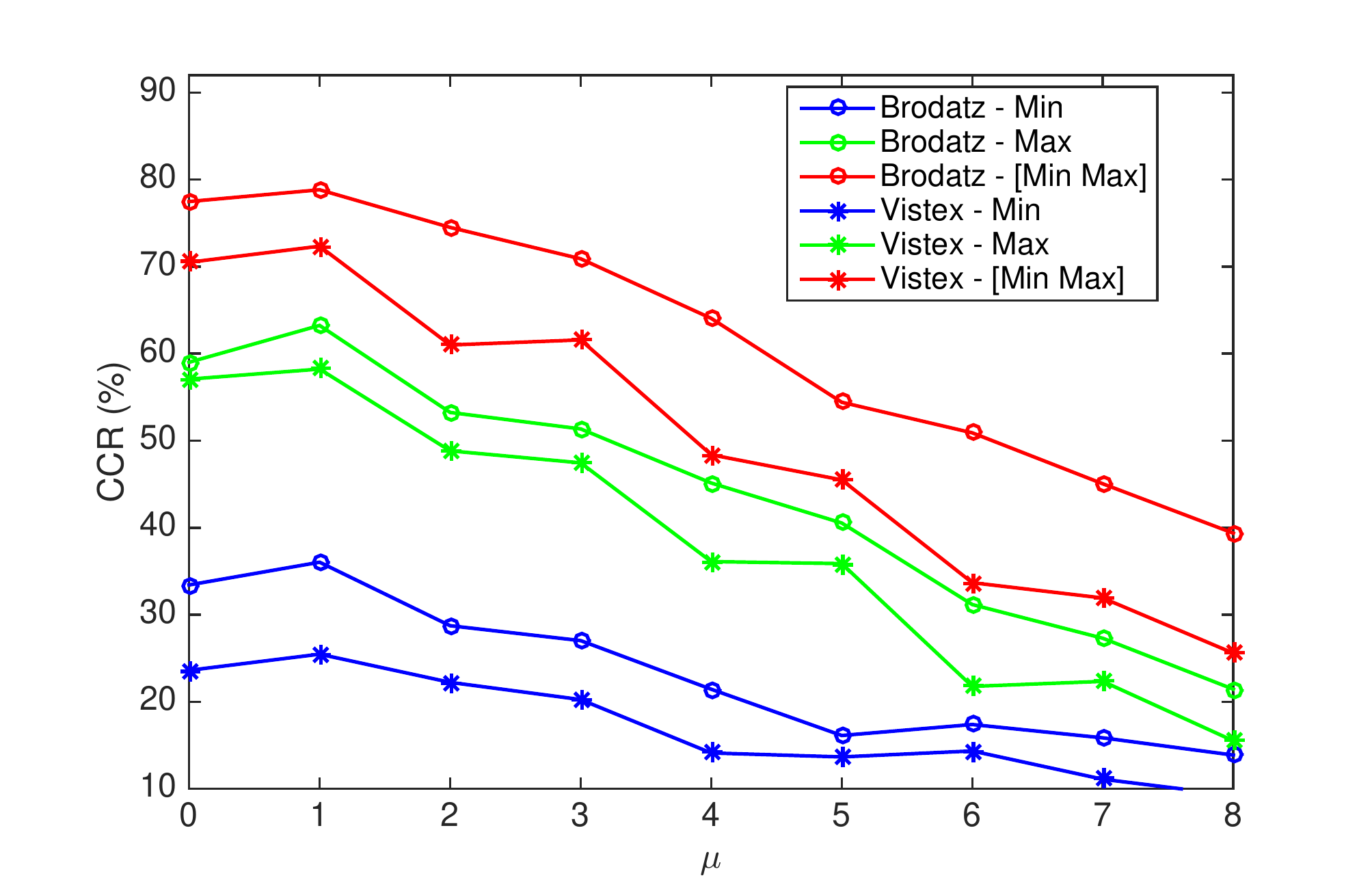}}\\
	\subfigure[Combining memory sizes $\mu$ ]{ \label{fig:memory_comb} \includegraphics[width=0.5\textwidth]{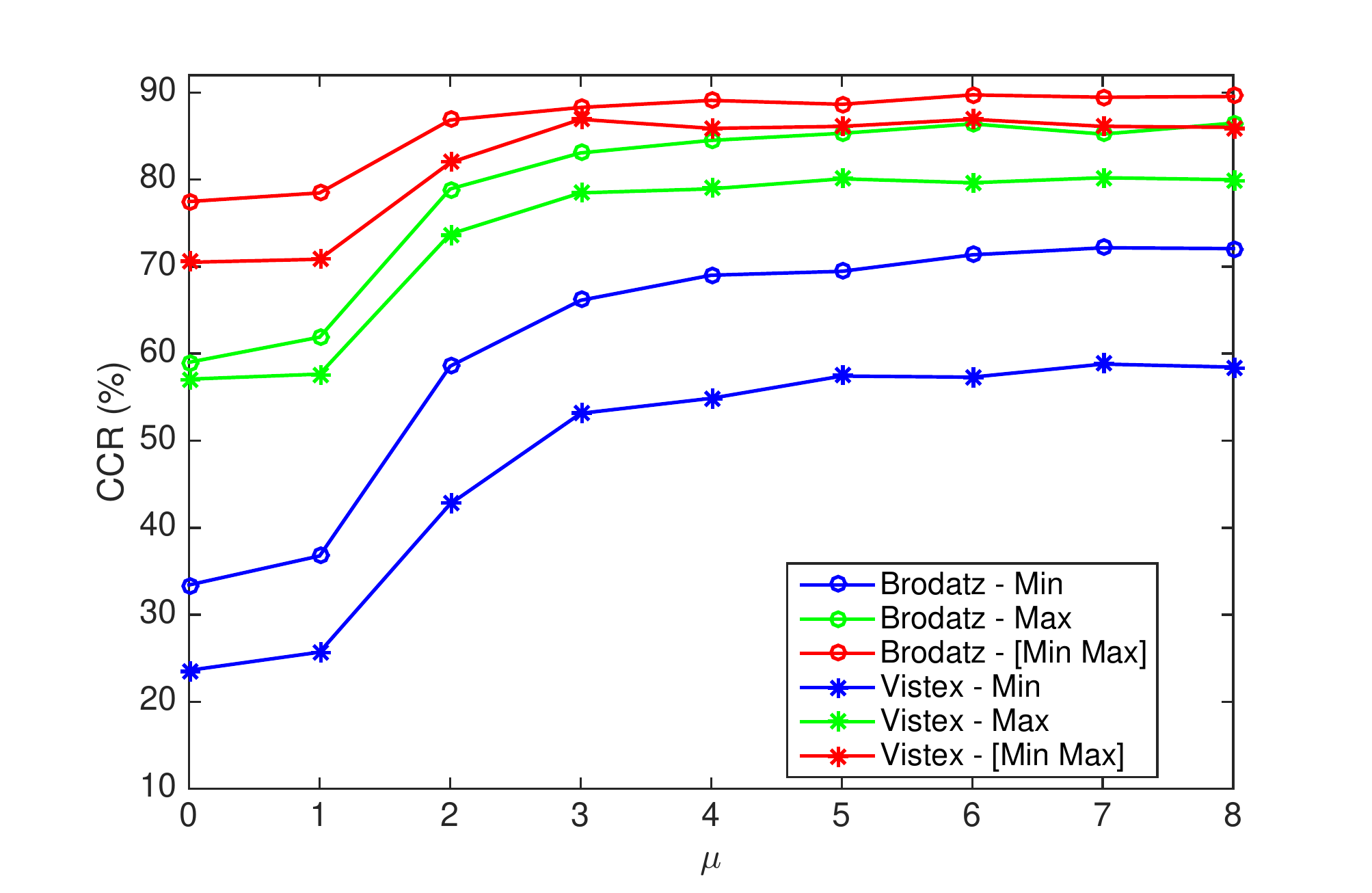}} 
	\caption{Correct classification rate as a function of different memory sizes $\mu$ and rule of movement $r$ .}
	\label{fig:ccr_u}
\end{figure}

In this section, we evaluated the performance of the deterministic tourist walk traditional on all image pixels using different parameters in the texture recognition task.
Figure \ref{fig:ccr_u} presents the correct classification rate (CCR) for different memory sizes $\mu$ and rules of movement $r$ used in the deterministic tourist walk. From Figure \ref{fig:memoryr}, we note that the correct classification rate decreases with the increase in the memory size $\mu$ on both Brodatz and Vistex datasets. The results in Figure \ref{fig:memoryr} also show that the tourist walk presents a better correct classification rate when uses the rule of movement $r=max$, instead of the  $r=max$. Therefore, tourist attractors formed in heterogeneous regions (i.e., regions with the presence of contours or changes in texture patterns) have most importance in recognition task \cite{backes2010texture}.  However, the best correct classification rate was achieved combining the rules $max$ and $min$ feature vectors ($\varphi^{min} \cup \varphi^{max}$). This new feature vector presents both heterogeneous and homogeneous image information. This occurs also on both Brodatz and Vistex datasets. 

Interesting results can be achieved concatenating the memory sizes. These results obtained for the combination of memory sizes are shown in Figure \ref{fig:memory_comb}. Note that the best correct classification rate is obtained for the concatenation of a few memory sizes. This diminishes the individual importance for each $\mu$ value. It is important to stress also that is necessary a small number of memory sizes provide better results. For the Brodatz and Vistex datasets, the concatenation of memory sizes that provided the best result is $\mu= [0, 1, 2, 3, 4, 5, 6]$, with a correct classification rate of 89.72\% and 86.92\%, respectively.

\subsection{Results}
In this section, we present the results of the proposed method for optimization of the deterministic tourist walk method by classifying the Brodatz and Vistex datasets.
We emphasize that our method uses exactly the same parameters obtained in the earlier experiment with the deterministic tourist walk on all image pixels ($\mu= [0, 1, 2, 3, 4, 5, 6]$ and $r =min,max$).

\begin{figure}[h!]
	\centering
	\includegraphics[width=.5\textwidth]{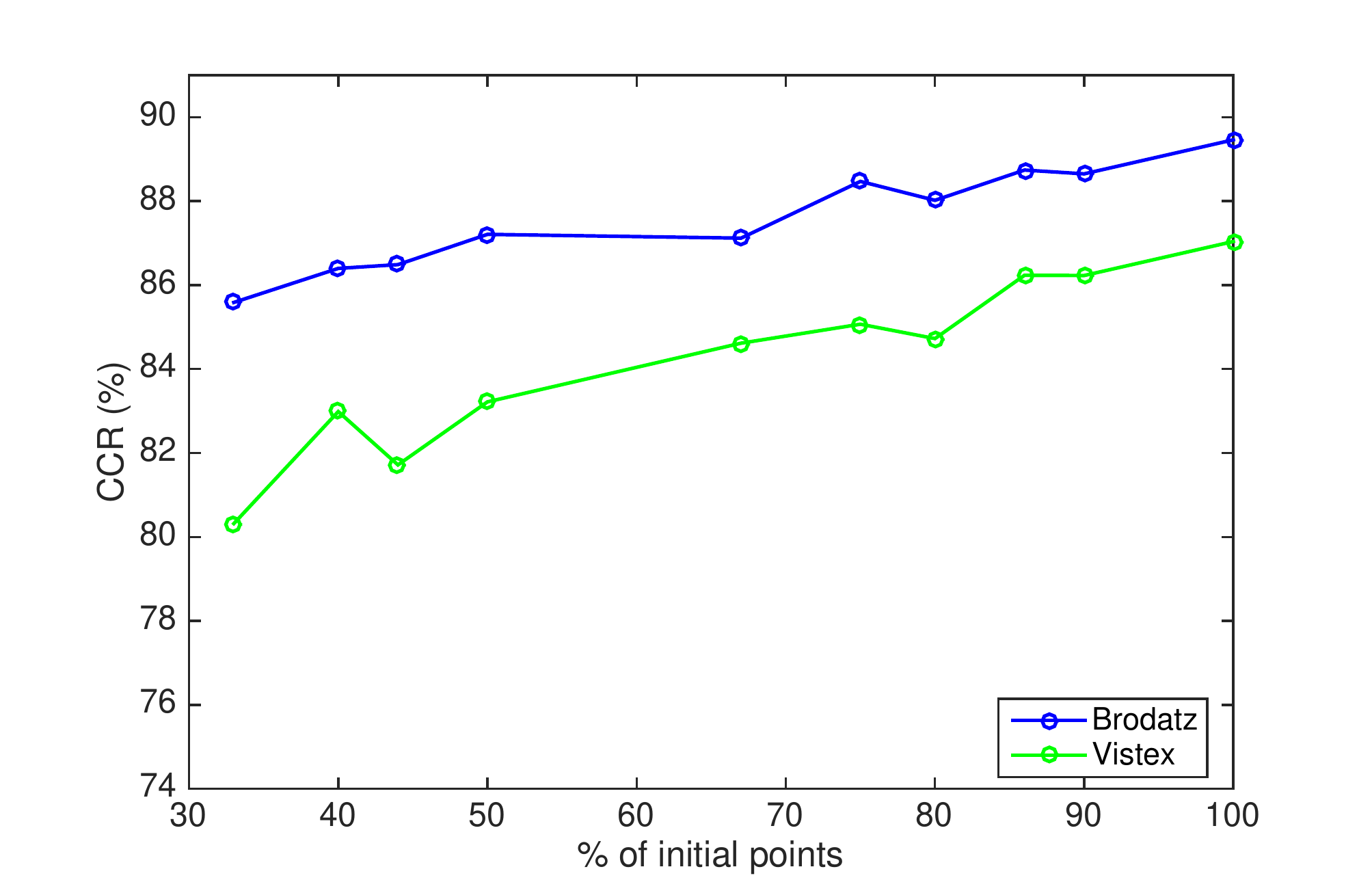}		
	\caption{Correct classification rate for Brodatz and Vistex datasets according to several \% of initial points.}
	\label{fig:results_k}
\end{figure}

First, we evaluated the performance of the fast deterministic tourist walk for different $k$ values according to Table \ref{tab:pct}.  
Results using different $k$ values from the Brodatz and Vistex datasets are shown in the plot of the Figure \ref{fig:results_k}.
In this Figure, we present the correct classification rate in function of the percentage of initial pixels which corresponding to a $k$ value. The plot shows that the good results were achieved for 50\% or less initial points. It is expected as, to decrease the percentage of initial points the correct classification rate also decreases in the same proportion. However, the results show that the correct classification rate decreases slowly. Notice that using 50\% of initial points the correct classification rate is 87.20 \% and 83.21 \% on Brodatz and Vistex datasets, respectively. In contrast, using all image pixels as initial points were achieved 89.45 \% and 87.03 \% on Brodatz and Vistex datasets, respectively. This corroborates our theory that is possible achieve good results in classification without to use all image pixels to perform the walk.

\begin{figure}[h!]
	\centering
	\includegraphics[width=.5\textwidth]{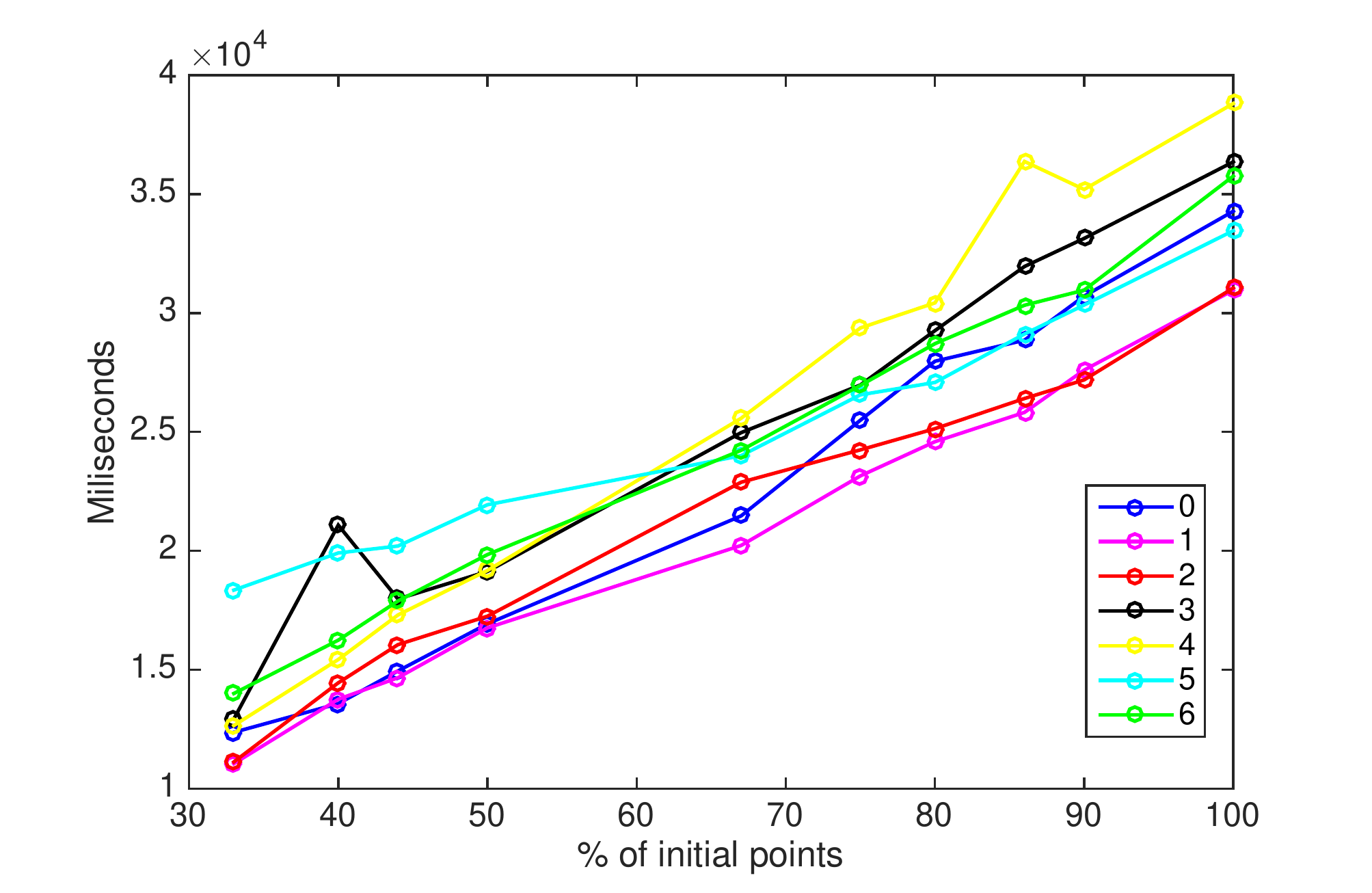}		
	\caption{Runtime of the fast deterministic tourist walk for each memory size and several \% of initial points from five images.}
	\label{fig:time}
\end{figure}

As we can see above, the correct classification rate is not highly affected by the decrease in the percentage of initial points of the walk.
Thus, it is necessary to answer other question: if the complexity computational also decreases significantly. 
For this, we evaluate the performance in the runtime from five images chosen randomly from Brodatz dataset. 
Figure \ref{fig:time} presents the runtime in milliseconds of the deterministic tourist walk on the five images for different percentage values. 
The experiments were performed using 2.6 GHz Intel (R) Core i5, 8 GB RAM and 64-bit Operating System.
In the figure, we show the runtime for all memory sizes used in the experiments.
It is observed that the runtime decreases sharply as the percentage of initial points decreases. This is verified for all memory sizes. 
These results indicate competitive runtime time combined with a good correct classification rate.

\section{Conclusion} \label{sec:conc}

In this paper, a study for optimization of the deterministic tourist walk has been presented. In this study, we propose a simple method that does not consider all image pixels as initial points of the walk. From this method, we evaluate the correct classification rate in function of the number of initial points to start a walk. Experimental results indicate that the use of a smaller number of initial points achieves a competitive correct classification rate. Besides that, the runtime is 
significantly lower than the use of all image pixels as initial points. For instance, in the Brodatz dataset to use 50\% of the pixels as initial points, the runtime decreases  for half, while correct classification rate decreases only 2.25\%.  
This result shows that our study can be applied for optimization of all methods that use the deterministic tourist walk for texture analysis.

\section*{Acknowledgment}

The authors gratefully acknowledge support from Coordination for the Improvement of Higher Education Personnel - CAPES (PROEX-9254772/M), CNPq (Grant Nos. 307797/2014-7 and 484312/2013-8) and FAPESP (Grant No. 14/08026-1).

\bibliographystyle{model1-num-names} 


\end{document}